\documentclass[a4paper]{article}

\usepackage{INTERSPEECH2019}
\usepackage[ruled]{algorithm2e}
\usepackage{graphicx,subcaption}
\usepackage{multirow}
\usepackage{amssymb,amsmath,amsthm,epsfig,bm}

\theoremstyle{definition}

\theoremstyle{theorem}

\theoremstyle{corollary}

\title{Acoustic Model Optimization Based On Evolutionary Stochastic Gradient Descent with Anchors for Automatic Speech Recognition}
\name{Xiaodong Cui and Michael Picheny}
\address{IBM Research AI \\ IBM T. J. Watson Research Center, Yorktown Heights, NY 10598, USA}
\email{\{cuix,picheny\}@us.ibm.com}

\begin{document}

\maketitle
\begin{abstract}
Evolutionary stochastic gradient descent (ESGD) was proposed as a population-based approach that combines the merits of gradient-aware and gradient-free optimization algorithms for superior overall optimization performance. In this paper we investigate a variant of ESGD for optimization of acoustic models for automatic speech recognition (ASR). In this variant, we assume the existence of a well-trained acoustic model and use it as an anchor in the parent population whose good ``gene" will propagate in the evolution to the offsprings. We propose an ESGD algorithm leveraging the anchor models such that it guarantees the best fitness of the population will never degrade from the anchor model. Experiments on 50-hour Broadcast News (BN50) and 300-hour Switchboard (SWB300) show that the ESGD with anchors can further improve the loss and ASR performance over the existing well-trained acoustic models.
\end{abstract}


\section{Introduction}

Evolutionary stochastic gradient descent (ESGD) was proposed in \cite{Cui_ESGD} for optimization of deep neural networks (DNNs). It is a population-based \cite{Jaderberg_PopNN} approach that integrates gradient-aware SGD and gradient-free evolutionary strategy (ES) \cite{Beyer_ESIntro}\cite{Lehman_ESNoJustFDA} in one framework to take advantage of the merits of both families of algorithms to deal with complicated optimization landscapes of DNNs.  In the meantime, a variety of SGD algorithms with various hyper-parameters are also taken into account in a co-evolution \cite{GP_COVNET}\cite{GP_COVEnsemble} setting of the population to leverage their complementariness. ESGD has shown its effectiveness in various domains such as speech, vision and language modeling.

The work to be reported in this paper is an extension of \cite{Cui_ESGD}.  In \cite{Cui_ESGD}, ESGD, given a fixed network architecture, starts optimization from scratch. In this paper, we investigate one variation of ESGD in which case we assume we already have some well-trained model in place to begin with. This is a fairly common scenario in acoustic modeling for ASR. For instance, we want to improvement performance over an existing baseline model. Since the well-trained existing model represents our prior knowledge with respect to acoustic modeling, we design an ESGD implementation to make use of it. Specifically, instead of the whole population starting from scratch, the existing model is used as an anchor inserted into the population as a parent so that its good ``genes" will quickly spread to the offsprings in the next generations to accelerate the evolution and hopefully give rise to improved solutions to the objective function. We design the ESGD with anchors in such a way that the best fitness of the new generation will not degrade from the previous generation. In other words, the best model in the new generation is no worse than that of the previous generation including the existing model in terms of fitness. Therefore, the proposed ESGD with anchors will guarantee a monotonic non-increase in loss function starting from the existing model.

We briefly investigated a prototype of ESGD using anchors in language modeling in \cite{Cui_ESGD} where the inserted anchor is never changed. In this work, we comprehensively investigate the anchor strategy on acoustic models. The contributions of this work include: (1) Design an ESGD algorithm with anchors to leverage the existing baseline models and switch anchors when better candidates surface; (2) Introduce an iterative process to apply ESGD with anchors to monotonically improve the best fitness of the population; (3) Conducted experiments on 50-hour Broadcast News (BN50) and 300-hour Switchboard (SWB300) to show the effectiveness of the proposed ESGD with anchor algorithm.

The remainder of the paper is organized as follows. Section \ref{sec:math} formulates the ESGD optimization problem mathematically. Its implementation with anchor models will be given in Section \ref{sec:esgdwa} along with the proof of its monotonic non-increasing best fitness of population. Section \ref{sec:exp} presents the experimental results on BN50 and SWB300 datasets followed by a discussion in Section \ref{sec:disc}. Finally, Section \ref{sec:sum} concludes the paper with a summary.

\section{Mathematical Formulation}
\label{sec:math}

Define the loss function
\begin{align}
  l_{i}(\theta) \triangleq \ell(h(x_{i};\theta),y_{i})
\end{align}
where $h$ is the function to be learned with parameter $\theta$ which maps the input space $\mathcal{X} \subseteq \mathbb{R}^{d_{x}}$ to the output space $\mathcal{Y} \subseteq  \mathbb{R}^{d_{y}}$ and $\{(x_{i},y_{i})\}_{i=1}^{n}\!\in\!\mathcal{X}\!\times\!\mathcal{Y}$. We further assume $\theta$ follows distribution $p(\theta)$, data follows distribution $p(\omega)$ and consider the expected empirical risk over $p(\theta)$ and $p(\omega)$
\begin{align}
    J = \mathbb{E}_{\theta}[\mathbb{E}_{\omega}[l_{\omega}(\theta)]] \label{eqn:esrisk}.
\end{align}
In ESGD \cite{Cui_ESGD}, $\omega\!\sim\!\text{Uniform}\{1, \cdots, n\}$ for SGD and a population of $\mu$ candidate solutions, $\{\theta_{j}\}_{j=1}^{\mu}$, is drawn to give the following average empirical risk of the population
\begin{align}
    J_{\mu} = \frac{1}{\mu} \sum_{j=1}^{\mu}\left(\frac{1}{n} \sum_{i=1}^{n} l_{i}(\theta_{j})\right) \label{eqn:esemprisk}
\end{align}
Eq.\ref{eqn:esemprisk} is the objective function under ESGD.

Following \cite{Cui_ESGD}, to avoid cluttered notation, we define the fitness function to be the empirical risk function, $\frac{1}{n} \sum_{i=1}^{n} l_{i}(\theta_{j})$, which is to be minimized. We define the $m$-elitist fitness: Let $\Psi_{\mu} = \{\theta_{1}, \cdots, \theta_{\mu}\}$ be a population with $\mu$ individuals and let $f$ be a fitness function associated with each individual in the population. Rank the individuals in the ascending order
\begin{align}
  f(\theta_{1:\mu}) \leq f(\theta_{2:\mu}) \leq \cdots \leq  f(\theta_{\mu:\mu})   \label{eqn:fsort}
\end{align}
where $\theta_{k:\mu}$ denotes the $k$-th best individual of the population \cite{Hansen_CMA}. We are interested in the fitness of the first $m$-best individuals $(1\!\leq\! m\! \leq\! \mu)$. When $m=1$, it amounts to the best fitness of the population.

Each individual in the population comes from a different optimizer with different hyper-parameters in each generation. After the gradient-based optimization, the individuals in the population will interact with each other in a gradient-free fashion to hopefully produce offsprings with better quality. In this way, the population-based ESGD not only can leverage the complementariness of various optimizers but also the complementariness of gradient-aware and gradient-free algorithms. Under a strategy of model back-off and elitist selection, ESGD guarantees that the best fitness in the population will never degrade (\cite{Cui_ESGD}, Corollary 1).

\section{ESGD with Anchors}
\label{sec:esgdwa}

Suppose we have a well-trained model in hand and want to further improve it without changing its architecture. Conventionally, we would pick an optimizer and its hyper-parameters and optimize the model until certain conditions (e.g. no improvement on the validation loss) met. This process is usually repeated multiple times and the best-performance model is chosen. ESGD can help this situation with a population of optimizers and let them interact in the evolution stage. Previously in \cite{Cui_ESGD}, ESGD starts with models with randomized weights without assuming the existence of any models. A baseline model reflects our prior knowledge in the model space, of which we can make use to also help ESGD. With this consideration, we propose the following ESGD algorithm with anchors (Algorithm \ref{alg:esgdwa}). The motivation behind this algorithm is that we always insert the best updated model, starting from the provided baseline model, into the population in the evolution as an anchor to make sure its ``genes" can spread out in future generations to not only accelerate the evolution process but also strive for better quality offspring.

\begin{algorithm}[!ht]
   \setlength{\textfloatsep}{0pt}
    \caption{ESGD with anchors}
    \label{alg:esgdwa}
     \textbf{Input:} generations $K$, \  SGD steps $K_{s}$, \ evolution steps $K_{v}$, \ parent population size $\mu$, \  offspring population size $\lambda$, elitist level $m$, initial anchor $\theta^{(0)}_a$ and anchor mating probability $p_{a}$.

     Initialize population $\mbox{\small$\Psi^{(0)}_{\mu} \leftarrow \{\theta^{(0)}_a\} \bigcup \{\theta^{(0)}_{1}, \cdots, \theta^{(0)}_{\mu-1}\}$}$;

     \tcp{$K$ generations}
     \For {$k = 1 : K$}
     {
         Update population $ \mbox{\small$\Psi^{(k)}_{\mu} \leftarrow  \Psi^{(k-1)}_{\mu}$} $;

         \vspace{0.2cm}
         \tcp{in parallel}
         \For {$j = 1 : \mu-1$}
         {
            Pick an optimizer $\mbox{\small$\pi^{(k)}_{j}$}$ for individual $\theta^{(k)}_{j}$;

            Select hyper-parameters of $\mbox{\small$\pi^{(k)}_{j}$}$ and set a learning schedule;

            \tcp{$K_{s}$ SGD steps}
            \For { $s = 1 : K_{s}$ }
            {
                SGD update of individual $\theta^{(k)}_{j}$ using $\mbox{\small$\pi^{(k)}_{j}$}$;

                If the fitness degrades, the individual backs off to the previous step $s\!-\!1$.
            }
         }

         \vspace{0.2cm}
         \tcp{$K_{v}$ evolution steps}
         \For { $v = 1 : K_{v}$}
         {
             Generate offspring population with anchor with probability $p_{a}$ and without anchor with probability $1-p_{a}$:
             $\mbox{\small$\Psi^{(k)}_{\lambda} \leftarrow \{\theta^{(k)}_{1}, \cdots, \theta^{(k)}_{\lambda}\}$}$;

             Sort the fitness of the parent (excluding anchor) and offspring population: $\mbox{\small$\Psi^{(k)}_{\mu+\lambda-1} \leftarrow \Psi^{(k)}_{\mu-1} \bigcup \Psi^{(k)}_{\lambda}$}$ ;

             Select the top $m$ ($m \leq \mu-1$) individuals with the best fitness ($m$-elitist);

             If the best fitness is better than the anchor, exchange the anchor and the current best individual;

             Update population $\mbox{\small$\Psi^{(k)}_{\mu}$}$ by combining $m$-elitist, randomly selected $\mu\!-\!m\!-\!1$ non-$m$-elitist candidates and updated anchor;
         }
     }
     Fine-tune the last parent population with a small learning rate using plain SGD for $K_{f}$ epochs. When fitness degrades, back off to previous epoch and anneal 2x learning rate;

     Pick the individual with the best fitness as the final model.
\end{algorithm}

In Algorithm \ref{alg:esgdwa}, the initial population is created by inserting the existing DNN model $\theta^{(0)}_{a}$ (the initial an anchor) into a family of otherwise randomly initialized networks. In the SGD step, the non-anchor models are optimized by a randomly selected SGD optimizer with randomly selected hyper-parameters for $K_{s}$ steps while the anchor model stays unchanged. The SGD optimization is implemented with model back-off. If the new fitness gets degraded, the individual will roll back to the previous step. In the evolution step, model encoding is first conducted where the weights of the models are vectorized into a real-valued vector. Then offsprings are produced following the $(\mu/\rho\!+\!\lambda)$-ES \cite{Beyer_ESIntro} rule:
\begin{align}
     \theta^{(k)}_{i} = \frac{1}{\rho}\sum_{j=1}^{\rho} \theta^{(k)}_{j} + \epsilon^{(k)}_{i}  \label{eqn:es_recomb_mutate}
\end{align}
where $\rho$ parents are randomly selected in proportional to their fitness and $\epsilon^{(k)}_{i} \sim \mathcal{N}(0,\sigma^{2}_{k})$ is Gaussian noise. Eq.\ref{eqn:es_recomb_mutate} amounts to a model average plus a perturbation. The probability of the anchor model showing up in Eq.\ref{eqn:es_recomb_mutate} is $p_{a}$. Therefore, about $p_{i}\lambda$ of the offsprings have the ``genes" from the anchor. After the offspring population is generated, their fitness are evaluated. If the best fitness is better than that of anchor, the model with the best fitness with switch with the anchor to become the new anchor for the next round. The top $m$ individuals are first selected and the rest $\mu\!-\!m\!-1\!$ individuals are then randomly picked from the non-$m$-elitist candidates. Finally, the updated anchor is inserted and new population is formed.

It can be shown the design of Algorithm \ref{alg:esgdwa} guarantees a non-degraded fitness from the initial anchor model. In fact, since the SGD step uses model back-off, the fitness of each individual in the population will not deteriorate. Furthermore, after the evolution step, the $m$-elitist strategy guarantees that individuals of the top $m$ fitness in the combined population of parent (excluding the anchor parent) and offspring population will enter the next generation. The best individual with the top-most fitness selected this way is then compared with the fitness of the anchor and whichever the better will become the next anchor. Therefore, the anchor in the next generation will be no worse than the anchor of the previous generation.

\section{Experiments}
\label{sec:exp}

Experiments are conducted on two datasets: BN50 and SWB300.

The 50-hour data in BN50 consists of a 45-hour training set and a 5-hour validation set.  The test set comprises 3 hours of audio. The acoustic models are fully-connected feed-forward network with 6 hidden layers and one softmax output layer with 5,000 states. There are 1,024 units in the first 5 hidden layers and 512 units in the last hidden layer. ReLU is used for the bottom 3 hidden layers and sigmoid for the rest. The input to the network is 9 frames of 40-dim LDA features derived from speaker-adapted PLP \cite{Hermansky_PLP} features.

The 300-hour data in SWB300 is split into 295 hours of training data and 5 hours of validation data. The test set is the Hub5 2000 evaluation set, composed of two parts: 2.1 hours of switchboard (SWB) data and 1.6 hours of call-home (CH) data. The acoustic models are bi-directional LSTM \cite{Hochreiter97} with 4 layers. Each layer contains 1,024 cells with 512 in each direction. On top of the LSTM layers, there is a linear bottleneck layer with 256 hidden units followed by a softmax output layer with 32,000 units.  The LSTMs are unrolled 21 frames. The input to the network is 140 in dimension which comprises 40-dim speaker-adapted PLP features after LDA and 100-dim i-vectors \cite{Dehak_ivec}).

The cross-entropy (CE) is chosen as the fitness function which is measured on the validation set. We consider various SGD variants and ADAM \cite{Kingma_adam} for the pool of optimizers. All optimization and fitness evaluation are carried out in parallel on multiple GPUs. We re-establish the single SGD baseline used in \cite{Cui_ESGD} as the references. The single baseline is trained using SGD with a batch size 128 without momentum for 20 epochs. The initial learning rate is 0.001 for BN50 and 0.025 for SWB300.  The learning rate is annealed by 2x every time the loss on the validation set of the current epoch is worse than the previous epoch and meanwhile the model is backed off to the previous epoch.

The size of the parent population is 100 and the offspring population 400. In ESGD, the hyper-parameters of optimizers which are to be chosen to create the parent population include batch size, learning rate, momentum and nesterov acceleration. Specifically, given a chosen optimizer, if applicable, there is a 80\% chance using the momentum and 20\% not using it.  When using the momentum, there is a 50\% chance using the nesterov acceleration. The momentum is randomly selected from $[0.1, 0.9]$. The batch size is uniformly chosen from $[64, 128, 256, 512]$. The learning rate is also randomly selected from a range $[a_{k}, b_{k}]$ depending on the generation $k$. The upper and lower bounds of the range are annealed over generations starting from the initial range $[a_{0}, b_{0}]$.  For ADAM, $\beta_{1}=0.9$ and $\beta_{2}=0.999$ are fixed and only the learning rate is randomized. The $m$-elitist strategy is applied to 60\% of the parent population. The probability of mating of the anchor model is 0.25. Table \ref{tab:hp} summarizes the hyper-parameter settings of the proposed ESGD with anchors for the two datasets.

\begin{table}[htb]
\caption{Hyper-parameters of ESGD with anchors for BN50 and SWB300}\label{tab:hp}
\centering
\begin{tabular}{l c c c c} \hline
\multicolumn{1}{c}{\multirow{2}{*}{\textbf{params}}}  &   \multicolumn{2}{c}{\textbf{BN50}}  &   \multicolumn{2}{c}{\textbf{SWB300}}    \\ \cline{2-5}
                    &  \textbf{SGD} &  \textbf{ADAM}  &  \textbf{SGD} &  \textbf{ADAM}      \\ \hline\hline
  $\mu$             &     100                 &     100                  &     100                    &     100       \\ \hline
  $\lambda$         &     400                 &     400                  &     400                    &     400       \\ \hline
  $\rho$            &     3                   &     3                    &     3                      &     3         \\ \hline
  $K$               &     15                  &     15                   &     15                     &     15        \\ \hline
  $K_{f}$           &     5                   &     5                    &     5                      &     5         \\ \hline
  $K_{s}$           &     1                   &     1                    &     1                      &     1         \\ \hline
  $K_{v}$           &     1                   &     1                    &     1                      &     1         \\ \hline
  $a_{0}$           &     1e-4                &     1e-4                 &     1e-2                   &     5e-5      \\ \hline
  $b_{0}$           &     2e-3                &     1e-3                 &     3e-2                   &     1e-3      \\ \hline
  $\gamma$          &     0.9                 &     0.9                  &     0.9                    &     0.9       \\ \hline
  $a_{k}$           &     $\gamma^{k}a_{0}$   &     $\gamma^{k}a_{0}$    &     $\gamma^{k}a_{0}$      &     $\gamma^{k}a_{0}$   \\ \hline
  $b_{k}$           &     $\gamma^{k}b_{0}$   &     $\gamma^{k}b_{0}$    &     $\gamma^{k}b_{0}$      &     $\gamma^{k}b_{0}$   \\ \hline
  momentum          &     [0.1, 0.9]          &     [0.1, 0.9]           &     [0.1, 0.9]             &     [0.1, 0.9]          \\ \hline
  $\sigma_{k}$      &     0.001               &     0.001                &     0.001                  &     0.001   \\ \hline
  $p_{a}$           &     0.25                &     0.25                 &     0.25                   &     0.25    \\ \hline
  $m$\_elitist      &     60\%                &     60\%                 &     60\%                   &     60\%    \\ \hline\hline
\end{tabular}
\end{table}

%\begin{table}[htb]
%\caption{The optimizers selected by the best candidate in the population over generations in BN50.}
%\label{tab:comp_opt}
%\centering
%\begin{tabular}{l|c|c} \hline
%      gen.        &    optimizer                              &    loss   \\ \hline\hline
%       1          &    ADAM, lr=1.73e-4, bs=64                &    2.443  \\ \hline
%       2          &    ADAM, lr=1.23e-4, bs=512               &    2.235  \\ \hline
%       3          &    SGD, lr=3.36e-4,nesv=F,mo=0.52,bs=512  &    2.100  \\ \hline
%       4          &    SGD, lr=4.05e-4,nesv=F,mo=0.45,bs=512  &    2.128  \\ \hline
%       5          &    SGD, lr=1.77e-4,nesv=F,mo=0.0,bs=64    &    2.035  \\ \hline
%       6          &    ADAM, lr=6.06e-5, bs=512               &    2.029  \\ \hline
%       7          &    SGD, lr=3.65e-4,nesv=F,mo=0.0,bs=512   &    2.059  \\ \hline
%       8          &    SGD, lr=6.05e-5,nesv=F,mo=0.37,bs=64   &    2.015  \\ \hline
%       9          &    SGD, lr=3.60e-4,nesv=F,mo=0.0,bs=128   &    1.992  \\ \hline
%       10         &    SGD, lr=7.48e-5,nesv=F,mo=0.38,bs=512  &    1.970  \\ \hline
%       11         &    ADAM, lr=5.96e-5, bs=256               &    1.968  \\ \hline
%       12         &    SGD, lr=2.45e-4,nesv=F,mo=0.0,bs=512   &    1.966  \\ \hline
%       13         &    ADAM, lr=4.68e-5, bs=256               &    1.955  \\ \hline
%       14         &    SGD, lr=7.52e-5,nesv=F,mo=0.0,bs=128   &    1.950  \\ \hline
%       15         &    SGD, lr=4.57e-5,nesv=F,mo=0.31,bs=128  &    1.946  \\ \hline
%\end{tabular}
%\end{table}

\begin{figure}[htb]
  \centering
  \centerline{\epsfig{figure=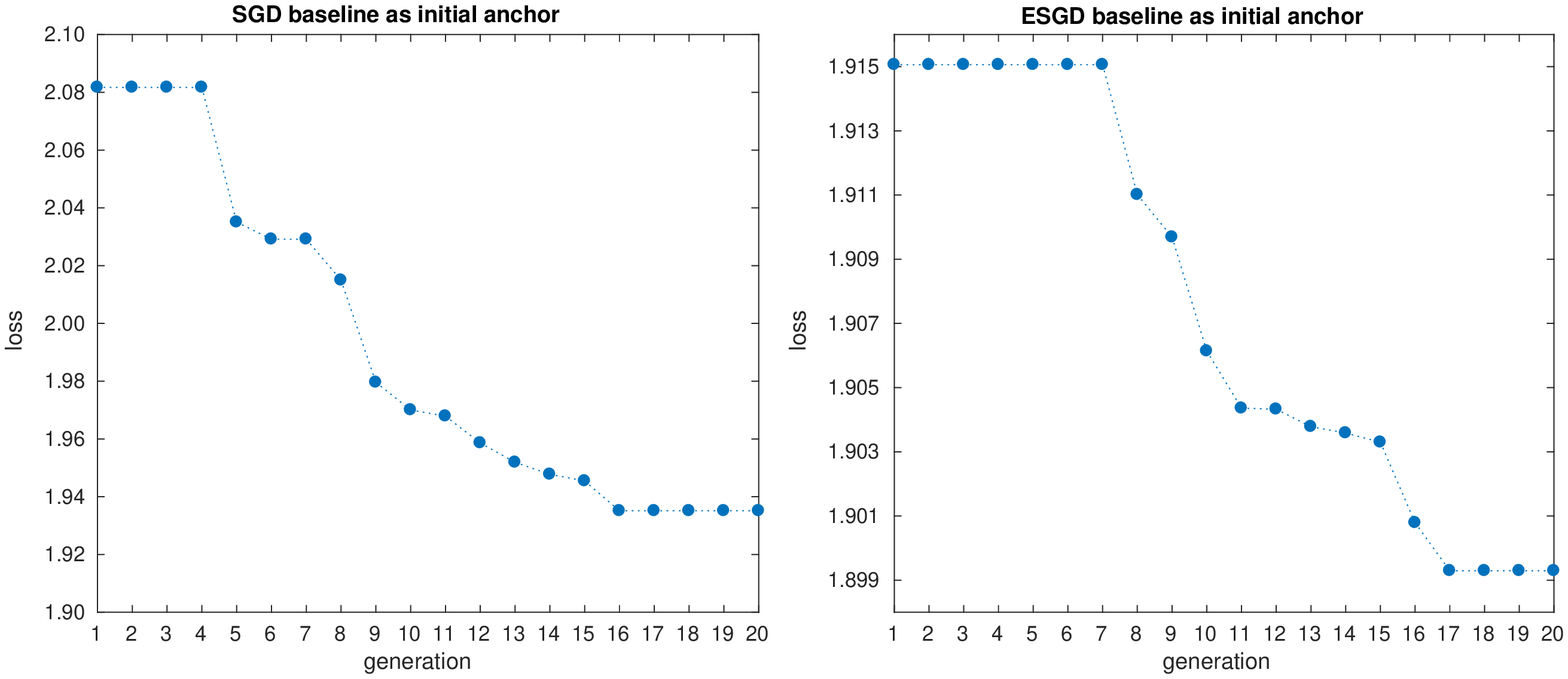, width=8cm, height=3.2cm}}
\caption{\label{fig:esgdwa}ESGD with anchors on BN50. Left panel shows the CE loss over generations using SGD baseline as initial anchor. Right panel shows the CE loss over generations using ESGD baseline as initial anchor.}
\end{figure}

Table \ref{tab:perf_bn50} shows the single SGD baseline (loss=2.082, WER=17.4\%, first row) and ESGD baseline (loss=1.916, WER=16.4\%, third row) on BN50 respectively. In the first experiments, we use the SGD baseline and ESGD baseline as the initial anchors. The evolution of the best fitness (minimal loss) of the population over generations is shown in the left and right panels of Fig.\ref{fig:esgdwa}, respectively. The drop of loss is monotonic. On the left panel, it takes 5 generations to break the SGD baseline while on the right it takes 8 generations. After 20 generations, ESGD using the SGD baseline as the initial anchor has the loss of 1.935 and WER 16.5, which improves significantly over the baseline model. Note that the SGD baseline is already a decent model trained under well-tuned training recipe. On the other hand, ESGD using the ESGD baseline as the initial anchor has the loss of 1.899 and WER 16.3, also improves from the ESGD baseline. The ESGD baseline is a very strong baseline. Using anchors is able to further improve it. These results are summarized in Table \ref{tab:perf_bn50}.

\begin{table}[htb]
\caption{CE loss and WER of ESGD with anchors on BN50}\label{tab:perf_bn50}\vspace{-0.5cm}
\centering
\begin{tabular}{l c c}  \\ \hline
 model                               &      loss        &     WER           \\ \hline\hline
 Single SGD baseline                 &     2.082        &    17.4           \\ \hline
 ESGD with anchor (SGD baseline)     &   \textbf{1.935} &  \textbf{16.5}    \\ \hline
% Population baseline                 &     2.029        &    17.1           \\ \hline
 ESGD baseline                       &     1.916        &    16.4           \\ \hline
 ESGD with anchor (ESGD baseline)    &   \textbf{1.899} &  \textbf{16.3}    \\ \hline
 ESGD with anchor (iterated anchor)  &   \textbf{1.882} &  \textbf{16.2}    \\ \hline
\end{tabular}
\end{table}

\begin{figure}[htb]
  \centering
  \centerline{\epsfig{figure=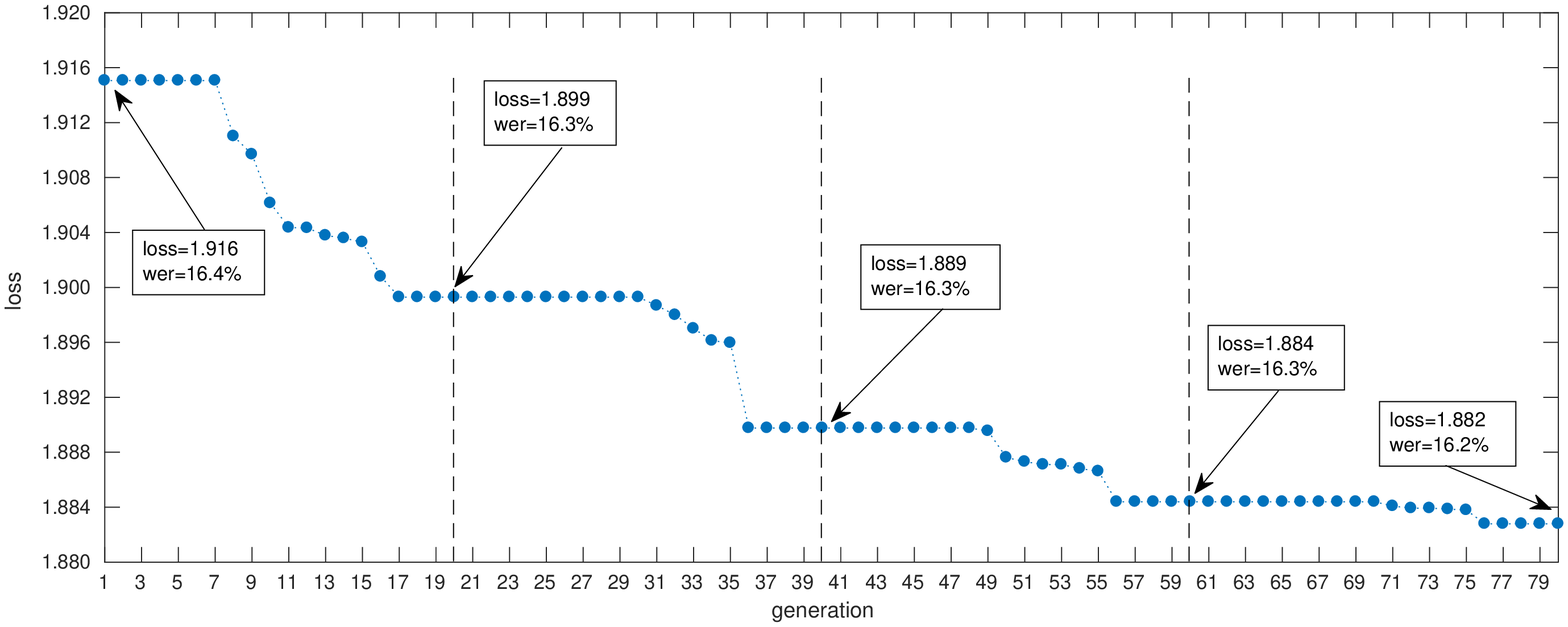, width=8cm, height=4cm}}
\caption{\label{fig:esgdwaiter}ESGD with iterative anchors on BN50.}
\end{figure}

The ESGD optimization with anchors can be conducted in an iterative fashion. After each round of evolution (in this case 20 generations), the best model can be used as the initial anchor model for the next round. The design of the ESGD will guarantee that the new anchor model will be no worse than the previous anchor in terms of loss. Therefore, the best model can be improved iteratively.  Fig.\ref{fig:esgdwaiter} demonstrates the evolution of the best loss of the population over 80 generations. After each 20 generations of evolution, the best model is used to update the anchor and restart the evolution process. As can be observed from the figure, the best loss drops monotonically and after 80 generations the loss is 1.882 and the WER is 16.2\%, which is shown in the last row of Table \ref{tab:perf_bn50}.

Table \ref{tab:perf_swb300} presents the experimental results of using ESGD with anchors on SWB300. The first two rows of the table provide the SGD and ESGD baselines, respectively, on SWB and CH test sets. When using ESGD baseline as the initial anchor, the CE loss improves from 1.551 to 1.536 after 20 generations. It slightly improves WER on CH (18.2\% $\rightarrow$ 18.1\%) but the WER on SWB stays the same (10.0\% $\rightarrow$ 10.0\%). If we start the ESGD all over with this best model, after another 40 generations with iterated anchors, the loss drops to 1.512 and the WER on CH further reduces to 17.7\% and the WER on SWB reduces to 9.9\%.

\begin{table}[htb]
\caption{CE loss and WER of ESGD with anchors on SWB300.}\label{tab:perf_swb300}
\centering
\begin{tabular}{lccc} \hline
 \multicolumn{1}{c}{\multirow{2}{*}{model}}    &  \multirow{2}{*}{loss}     &    \multicolumn{2}{c}{WER}    \\ \cline{3-4}
                                               &                            &    SWB        &        CH      \\ \hline\hline
 Single SGD baseline                 &     1.648        &    10.4         &    18.5         \\ \hline
 ESGD baseline                       &     1.551        &    10.0         &    18.2         \\ \hline
 ESGD with anchor (ESGD baseline)    &   \textbf{1.536} &  \textbf{10.0}  &  \textbf{18.1}  \\ \hline
 ESGD with anchor (iterated anchor)  &   \textbf{1.512} &  \textbf{9.9}   &  \textbf{17.7}  \\ \hline\hline
\end{tabular}
\end{table}

\section{Discussion}
\label{sec:disc}

Parallel computing is a necessity for ESGD which is a powerful approach when there is strong computational power in hand. The reported experiments are carried out in a distributed manner where SGD and fitness evaluation are conducted on multiple GPUs in parallel, the number of which is roughly the number of individuals in the parent population ($\sim$100). Under this condition, the wall clock time of ESGD is about the same as that an end-to-end vanilla SGD run.

A healthy population diversity is crucial for good performance of ESGD. It can prevent pre-mature convergence in ES and give good chances to produce better offsprings. We ensure this by using complementary optimizers and employing very short SGD and ES updates in each ESGD generation ($K_{s}=K_{v}=1$). Moreover, the initial parent population is created using models with randomized weights. This seems to be hurtful in the sense of fitness but in the long run it helps to establish the diversity in the population and turns out to be better than just slightly perturb the initial anchor model to create the population.

Anchor models can help to accelerate the evolution process as the good ``genes" of an anchor can spread out (with probability) to the next generations until it is replaced by another anchor with better ``genes". This is the motivation behind the design of the proposed algorithm. With model backoff, elitist and anchor switching, it is guaranteed that the fitness of the best model of the population will never degrade. This has been demonstrated in Figs.\ref{fig:esgdwa} and \ref{fig:esgdwaiter}. In the worst case, after the evolution no better model is found and the initial anchor survives as the best model.

The objective to optimize under ESGD is fitness which is CE loss in this work. In essence, there is no difference from other deep learning optimization problems except that it is population based and makes use of complementary optimizers and gradient-aware/gradient-free algorithms.  From this perspective, it also has to cope with the training vs. generalization issue. We evaluate the fitness on the validation set, which is one way to help generalization. But observations are also made that there are cases where better fitness may not lead to better generalization (e.g. a narrow energy well in the landscape can give a low loss but may not generalize well \cite{Choromanska_LossSurf}\cite{Chaudhari_EnSGD}). Furthermore, although CE is correlated with WER, they are not exactly aligned. A better CE loss may not give rise to better WER. Looking forward, we hope to explore alternative fitness functions such as sequence training criteria sMBR \cite{Kingsbury_Seqtrain} and MPE \cite{Povey_MPE} which are better aligned with WER, the ultimate objective of ASR. Regularized fitness functions which can help generalization are another direction worth exploring.

\section{Summary}
\label{sec:sum}

In this paper, we investigated a population-based ESGD algorithm assuming some well-trained model exists. We use this model as an anchor in the population to accelerate the evolution and improve the quality of offsprings. We introduced anchor switching in the population and also an iterative way of applying ESGD with anchors to monotonically improve the best fitness of the population. We carried out experiments on BN50 and SWB300 and demonstrated the monotonic decrease of the CE loss. In some cases, it will also help improve the WERs over strong ESGD baselines.

\bibliographystyle{IEEEtran}

\bibliography{esgdwa}

\end{document}